# A framework of motion capture system based human behaviours simulation for ergonomic analysis

Ruina MA, Damien CHABLAT , Fouad BENNIS, Liang MA

IRCCyN / Ecole Centrale de Nantes, 1 Rue de la Noë, 44321 Nantes Cedex 3, France
{ruina.ma, damien.chablat, fouad.bennis}@irccyn.ec-nantes.fr, liangma@tsinghua.edu.cn

## Introduction

With the increasing of computer capabilities, Computer aided ergonomics (CAE) offers new possibilities to integrate conventional ergonomic knowledge and to develop new methods into the work design process. As mentioned in [1], different approaches have been developed to enhance the efficiency of the ergonomic evaluation. Ergonomic expert systems, ergonomic oriented information systems, numerical models of human, etc. have been implemented in numerical ergonomic software. Until now, there are ergonomic software tools available, such as Jack, Ergoman, Delmia Human, 3DSSPP, and Santos, etc. [2-4]. The main functions of these tools are posture analysis and posture prediction. In the visualization part, Jack and 3DSSPP produce results to visualize virtual human tasks in 3-dimensional, but without realistic physical properties. Nowadays, with the development of computer technology, the simulation of physical world is paid more attention. Physical engines [5] are used more and more in computer game (CG) field. The advantage of physical engine is the nature physical world environment simulation. The purpose of our research is to use the CG technology to create a virtual environment with physical properties for ergonomic analysis of virtual human.

In the following sections, firstly we will introduce motion capture system and the related data obtained and processed. Secondly we will illustrate the physical engine, and by using it we create our virtual physical environment. Thirdly, we propose a simulation framework using motion capture system and physical engine for ergonomic analysis. At last, we introduce a study case to present the ergonomic analysis based on a muscle fatigue model.

## Motion capture

Motion capture techniques have been applied frequently to obtain the dynamic and natural motion information in human simulation tools [6]. Several kinds of tracking systems are available, such as mechanical motion tracking, acoustic tracking, magnetic tracking, optical motion tracking, and inertial motion tracking [7]. We use the motion capture system which is produced by NaturalPoint Company [8] to obtain the trajectory of human movement.

**Optical tracking system camera placement:** Our optical capture system is composed of three cameras around the work space. Markers are made from reflective material that could be tracked by camera sensitively, and they are attached to the human articulation. In order to track markers, multiple OptiTrack cameras must be arranged to have several overlapped fields of view. For best calibration and tracking results, it should be avoided to place al the cameras in the same plane, instead position the cameras at different angles. This will create an area called a capture volume in which tracking can occur. The overall capture system works at the frequency 100 Hz. It could offer a minimum sampling rate to have good image rendering, and this could provide sufficient detailed analysis of the static and dynamic human body motions.

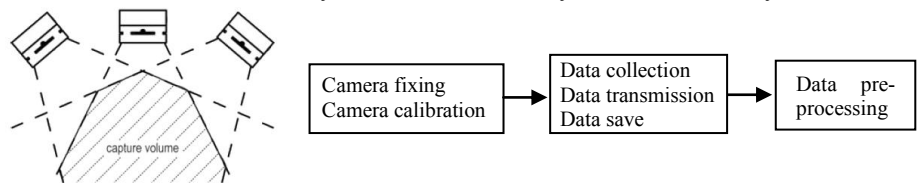

**Fig. 1.** Camera Installation and Data Obtain Process

**Data obtain process:** Camera calibration here refers to the calibration of the multi-camera system. It is the base of the following work. Dynamic camera calibration method which offered by the software was used to calibrate multiple cameras. In the Data Collection part, we open the interface about data collection. Markers are placed in the proper articulation, in order to get the relatively accurate position. In each articulation we put at least two markers and organise them as a tracker table. In this situation the average spatial position values of the markers will be the final trajectory values. In order to get the high quality trajectory , these markers moving in the capture volume space, should be viewed by at least two cameras.

The data we get from the software is the format ".cvs". In this format it contains much information more than we want. We must extract the information that we are interested in. In the data pre-processing part we do the data extraction using Matlab.

## Physical engine

A physics engine is computer software that provides an approximate simulation of certain simple physical systems, such as rigid body dynamics (including collision detection), soft body dynamics, and fluid dynamics. They are use in films, computer graphics, and mainly in video games (typically as middleware), in which the simulations are in real-time. The term is sometimes used more generally to describe any software system for simulating physical phenomena, such as high-performance scientific simulation [9]. For the human being, the dynamic motion is related more about the mechanics. Doing analysis in the physics world is more meaningful.

**Physical engine development:** A physics engine is basically like a black box; it uses a set of mathematical expressions to simulate physics. Data drives the physics engine to develop more quickly. There are two compelling advantages to use a physics engine in the simulation. The first one is time saving and the second one is quality of rendering. A physical engine provides us with the ability to have interactive effects in believable ways. Using it we could create a more realistic virtual world.

**Software Design:** Bullet has been designed to be customizable and modular. The main components are organized as follows:

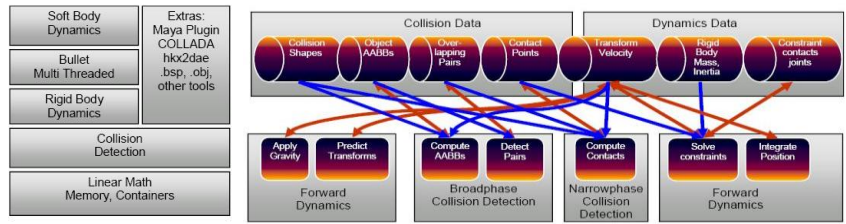

**Fig. 2.** Bullet Engine Software Design Structure [10]

**Rigid Body Physics Pipeline:** Before going into details, the following diagram shows the most important data structures and computation stages in the Bullet physics pipeline. This pipeline is executed from left to right, starting by applying gravity, and ending by position integration, updating the world transform.

The entire physics pipeline computation and its data structures are represented in Bullet by a dynamics world. When performing 'stepSimulation' on the dynamics world, all the above stages are executed. The default dynamics world implementation is the 'btDiscreteDynamicsWorld'.

Finally, based on the framework of Bullet, we created a virtual environment with physic and robotic properties. According to the digital human structure, we create our mannequin in the environment.

## Framework of simulation

The framework presented in Fig. 3 illustrating the workflow of the software system which was consisted by mainly four parts: The first part is using motion capture system to get the trajectory information of human body movements. The second part is using digital human model to get related geometric data and dynamic information data. The forth part is using physical engine to create a physical virtual environment and create the virtual human in the environment. The forth part is immerge the dynamic muscle fatigue model when the human doing the movement in order to do the ergonomic analysis.

We use the physical engine to create the virtual physical environment and to create the virtual human in this environment. We have read the data obtained from the motion capture system and put it in a virtual human in the virtual physical environment. We had set up the dynamic muscle fatigue model, and immerge it in the virtual human is the next step work.

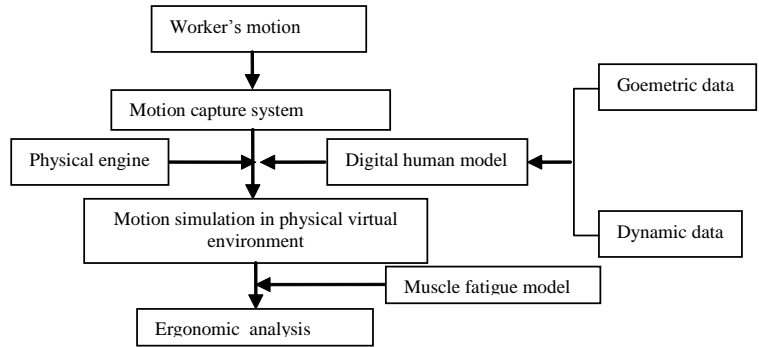

**Fig. 3.** Simulation Framework

## Case study

Considering the complexity of the whole body movement, at the very beginning we set the posture of an arm in moving an object as the research target.

**Arm geometric model:** We put three groups of markers on the arm articulation position (the shoulder, the elbow and the wrist). In order to reduce the complexity and avoid the redundant situation, we suppose in the shoulder there is two degree of freedom, and in the elbow there is one degree of freedom and the wrist is fixed. The geometric model is as below:

| j | σ(j) | α (j) | D(j) | θ(j) | R(j) |
|---|---|---|---|---|---|
| 1 | 0 | π/2 | 0 | $\theta_1$ | 0 |
| 2 | 0 | - π /2 | 0 | $\theta_2$ | 0 |
| 3 | 0 | 0 | $D_3$ | $\theta_3$ | 0 |
| 4 | 0 | 0 | $D_4$ | 0 | 0 |

**Tab. 1.** Geometry Parameters of Arm   [11]

$D_3$ and $D_4$ represent the length of upper and lower arm. Using the trajectory which we get from motion capture system and with the inverse kinematic model, we could calculate the articulation angle $\theta_1, \theta_2, \theta_3$.

**Ergonomic analysis---muscle fatigue model:** Ergonomic analysis and assessment can be evaluated from different aspects according to the task requirement. At first, most of the ergonomic analysis is based on the scoring method. With the development of the ergonomic field, there come out some model to describe the working situations. Currently, we pay attention to the muscle fatigue effectors in ergonomic analysis. Recently, we have defined a muscle fatigue model **[12]** from the macroscopic point of view and using simple parameters to define the muscle fatigue development when time elapses. We want to integrate the fatigue model in the human moving procedure. Below is the figure to illustrate the muscle fatigue trend with time goes by.

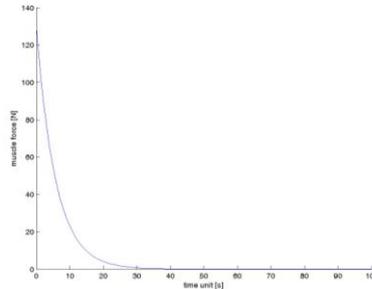

**Fig. 4.** Muscle Fatigue Trend with Time  [12]

## Conclusion and perspective

In this paper, we proposed a framework for human behavior simulation. We use physical engine to set up the virtual environment and the purpose of this framework is to enhance the ergonomic analysis. Ergonomic analysis is important challenge but we have concentrate our research on muscle fatigue model to use this model into the framework for trajectory planning is our final purpose. This is only a preliminary result. Thus, there still a many research to do in the ergonomic module and visualization module. Later we will take consideration of path planning and collision avoiding based on a multi-agent optimization approach in the physical world.